\documentclass[a4paper,12pt]{article}
\usepackage{epsf}
\usepackage{epsfig}
\usepackage{graphics}
\usepackage{graphicx}
\usepackage{times,amsfonts,amssymb,amsthm,amsmath,delarray,dcolumn}
\usepackage{array}
\usepackage{supertabular} 
\usepackage{longtable}
\usepackage{tabularx}
\usepackage{natbib}
\newif\ifpdf 
    \ifx\pdfoutput\undefined 
    \pdffalse 
    \else 
	\pdfoutput=1 
\usepackage[colorlinks,bookmarks,pdftitle={Weimar Optimization and Stochastic Days},pdfsubject={WOSD},pdfkeywords={Stochastic,Optimization}, pdfauthor={No name}]{hyperref}
    \pdftrue 
    \fi
\usepackage{fancyhdr}
\title{Recent advances in Meta-model of Optimal Prognosis}
\author{Thomas Most$^{1}$\thanks{Contact:
Dr.-Ing. Thomas Most, DYNARDO -- Dynamic Software and Engineering GmbH, 
Luthergasse 1d,
D-99423 Weimar, Germany, E-Mail: thomas.most@dynardo.de}
 \& Johannes Will$^{1}$ \\ \\
{\scriptsize $^1$ DYNARDO -- Dynamic Software and Engineering GmbH, Weimar,
Germany}}
\date {}
\begin{document}
\maketitle
\begin{abstract}
In real case applications within the virtual prototyping process, it is not
always possible to reduce the complexity of the physical models
and to obtain numerical models which can be solved quickly. Usually, every
single numerical simulation takes hours or even days. Although the
progresses in numerical methods and high performance computing, in such cases,
it is not possible to explore various model configurations, hence efficient 
surrogate models are required.

Generally the available meta-model techniques show several advantages and disadvantages
depending on the investigated problem. In this paper we present an automatic approach 
for the selection of the optimal suitable meta-model for the actual problem.
Together with an automatic reduction of the variable space using advanced filter techniques
an efficient approximation is enabled also for high dimensional problems.

\end{abstract}
{\bf Keywords:} surrogate models, meta-modeling, regression analysis, 
optimal prognosis, variable reduction

\newpage

\section{Introduction}
Meta-modeling
is one of the most 
popular strategy for design exploration within 
nonlinear optimization and stochastic analysis (see e.g.\ \cite{Booker1999,Giunta1998,Simpson2003}). Moreover, the engineer has to 
calculate the general trend of physical phenomena or would like to re-use design
experience on different projects.
Due to the inherent 
complexity of many 
engineering problems it is quite alluring to approximate 
the problem and to solve other design configurations in a smooth sub-domain by 
applying a surrogate model (\cite{Sacks1989,Simpson2001}). Starting from a reduced number of simulations, 
a surrogate model of the original physical problem can be used to perform
various possible design configurations without computing any further analyzes.
In one of our previous publications (\cite{Roos_2007_WOST}) we investigated several meta-model types and variable reduction techniques
by means of several examples. In this previous paper we summarized that no universal approach exists and the optimal filter configurations
can not be chosen generally. Therefor we developed an automatic approach for this purpose based on a library of available meta-models
and tools for variable reduction. This approach serve us based on a new measure for the approximation quality the Meta-model of Optimal Prognosis.
This optimal meta-model serves the best compromise between 
  available information (samples) and model representation in terms of considered input variables.

This paper is constructed as follows: first we present the meta-model approaches which are used later in our investigations.
Then we investigate different measures which are used to assess the prognosis quality of the approximation model.
In the fourth section we present our procedure to estimate global parameter sensitivities in the reduced parameter space of the optimal meta-model.
Finally we present the framework of the meta-model selection and its implementation in optiSLang.

\section{Meta-model approaches}
\subsection{Polynomial regression}
\begin{figure}[th]
	\begin{minipage}[t]{.48\linewidth}
	\centerline{\includegraphics[width=\textwidth]{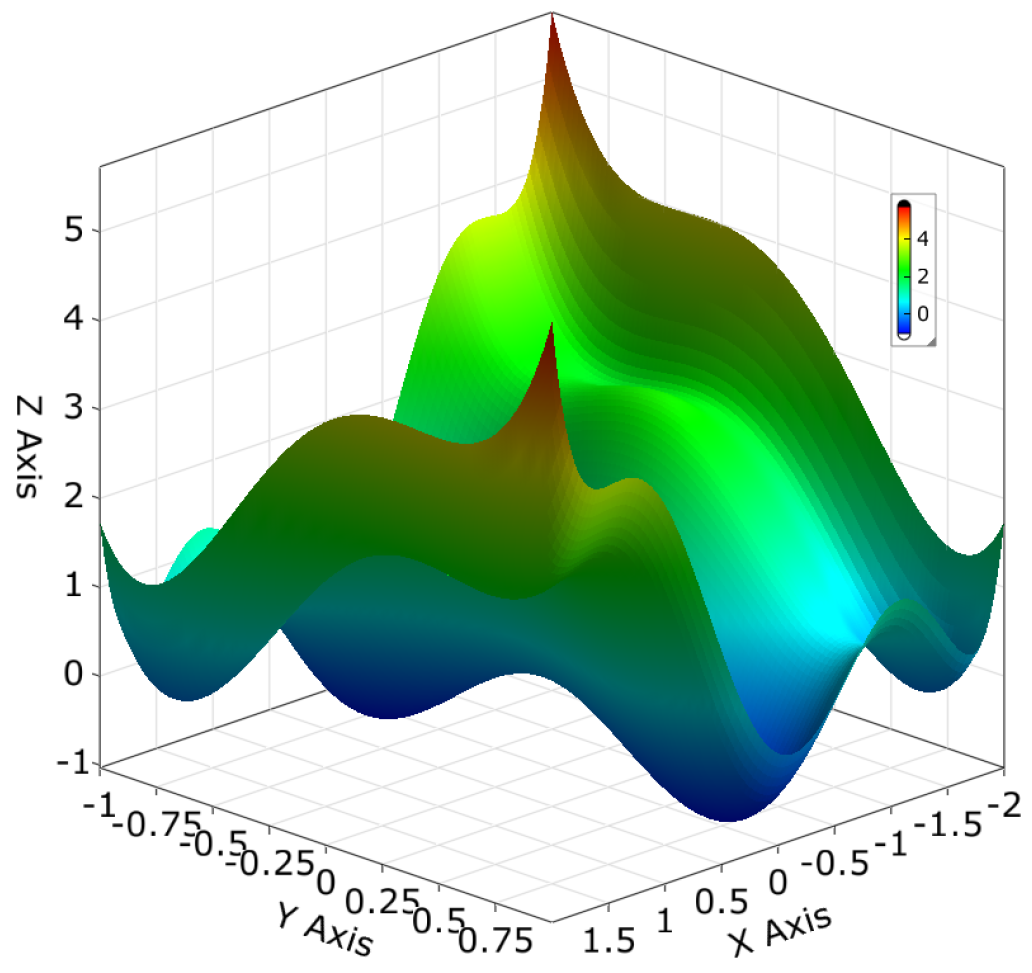}}
  		\caption{Original model response function $z(x,y)$}
   		\label{original}
	\end{minipage}\hfill
	\begin{minipage}[t]{.48\linewidth}
	\centerline{\includegraphics[width=\textwidth]{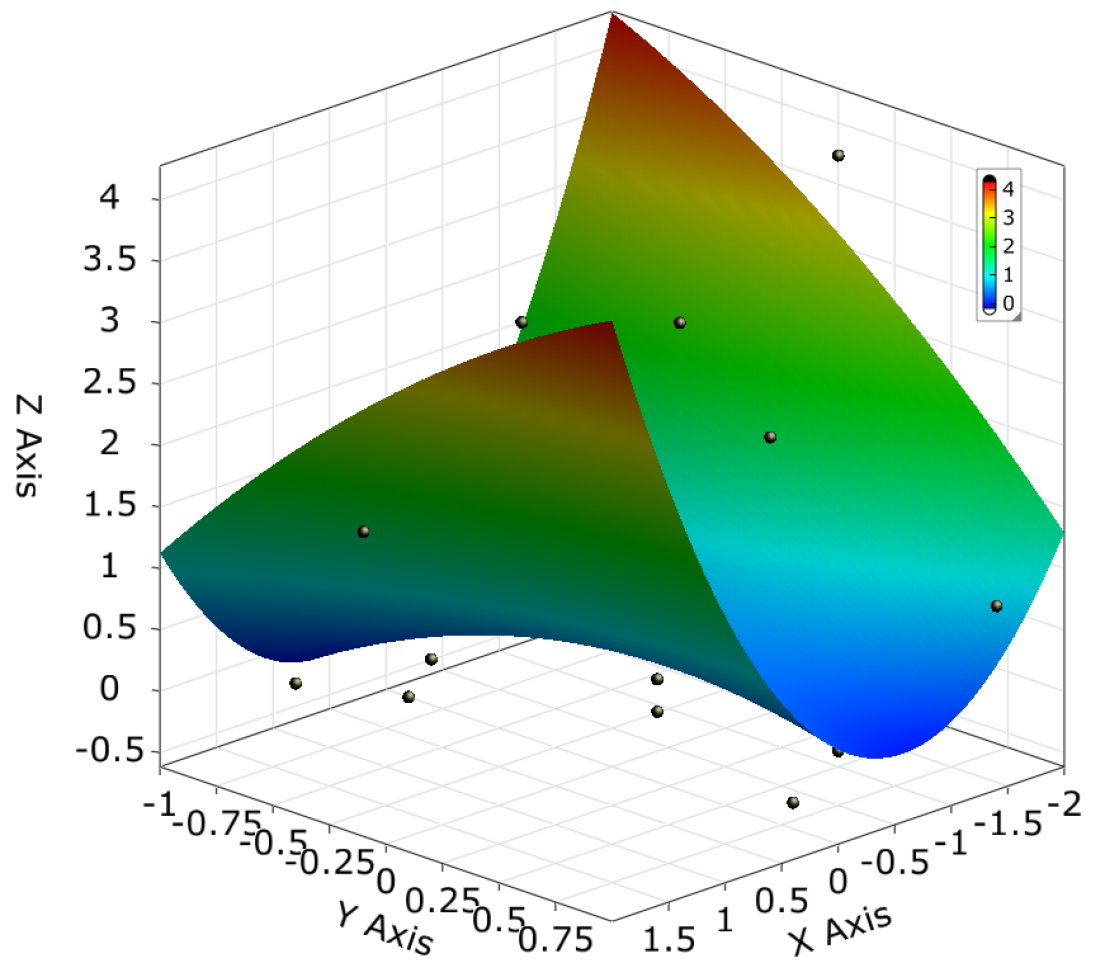}} 
  		\caption{Polynomial least square approximation with quadratic basis}
   		\label{approx_quadratic_2}
	\end{minipage}
\end{figure}

A commonly used approximation method is polynomial regression,
where the model response is generally approximated by a polynomial basis function
\begin{equation}
\begin{aligned}
\mathbf{p}^T(\mathbf{x})=&\left[ 1\;x_1\;x_2\;x_3\; \ldots \;x_1^2\;x_2^2\;x_3^2\; \ldots \right.\\
&\quad\left.\ldots\;x_1x_2\;x_1x_3\;\ldots\; x_2x_3\;\ldots\right]
\end{aligned}
\end{equation}
of linear or quadratic order with or without linear coupling terms.
The model output $y_j$ for a given sample $\mathbf{x}_j$ of the input parameters $\mathbf{X}$
can be formulated as the sum of the approximated value $\hat y_j$ and an error term $\epsilon_j$
\begin{equation}
y(\mathbf{x}_j) = \hat y_j(\mathbf{x}_j) + \epsilon_j = \mathbf{p}^T(\mathbf{x}_j)\boldsymbol{\beta} + \epsilon_j,
\end{equation}
where $\boldsymbol{\beta}$ is a vector containing the unknown regression coefficients.
These coefficients are generally estimated from a given set of
sampled support points by
assuming independent errors with equal variance at each point.
By using a matrix notation the resulting least squares solution reads
\begin{equation}
\boldsymbol{\hat \beta}=(\mathbf{P}^T\mathbf{P})^{-1}\mathbf{P}^T \mathbf{y}
\end{equation}
where $\mathbf{P}$ is a matrix containing the basis polynomials of the support point samples.

Since in the MOP approach the original parameter
space is reduced to a much smaller subspace of important variables, small artificial noise has to be smoothed by the approximation
method. Polynomial regression is very suitable for this purpose. Nevertheless, only global linear or quadratic 
function can be represented. For more complex coherence between model input and output
local regression methods have to be used instead of global polynomial regression.

\subsection{Moving Least Squares approximation}
\begin{figure}[th]
	\begin{minipage}[t]{.48\linewidth}
	\centerline{\includegraphics[width=1.15\textwidth]{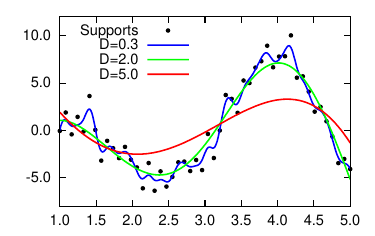}}
  		\caption{Classical MLS approximation depending on the influence radius $D$}
   		\label{MLS_classic}
	\end{minipage}\hfill
	\begin{minipage}[t]{.48\linewidth}
	\centerline{\includegraphics[width=1.15\textwidth]{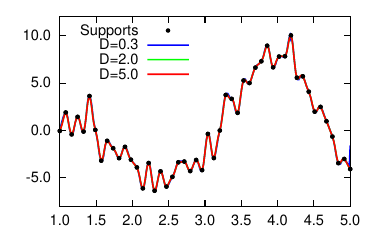}} 
  		\caption{Interpolating MLS approximation with regularized weighting}
   		\label{MLS_reg}
	\end{minipage}
\end{figure}

In the Moving Least Squares approximation a local character of the regression
is obtained by introducing 
position depending radial weighting functions.
MLS approximation can be understood as an extension of the polynomial regression.
Similarly the basis function can contain every type of function, but generally only linear and quadratic terms are used.
This basis function can be represented exactly by obtaining the best local fit for the actual interpolation point.
The approximation function is defined as
\begin{equation}
\hat y(\mathbf x)=\mathbf{p}^T(\mathbf{x})\mathbf{a}(\mathbf{x})
\end{equation}
with changing (``moving'') coefficients $\mathbf{a}(\mathbf{x})$ in contrast to the global coefficients of the polynomial regression.
The final approximation function reads
\begin{equation}
\hat y (\mathbf{x})=\mathbf{p}^T(\mathbf{x})(\mathbf{P}^T\mathbf{W(x)}\mathbf{P})^{-1}\mathbf{P}^T \mathbf{W(x)} \mathbf{y}
\end{equation}
where the diagonal matrix $\mathbf{W(x)}$ contains the weighting function values corresponding to each support point.
Distance depending weighting functions $w=w(\| \mathbf{x}-\mathbf{x}_i\|)$ have been introduced.
Mostly the well known Gaussian weighting function is used
\begin{equation}
w_{exp}(\| \mathbf{x}-\mathbf{x}_i\|) = exp\left(-\frac{\| \mathbf{x}-\mathbf{x}_i\|^2}{\alpha^2 D^2}\right)
\end{equation}
where the definition of the influence radius $D$ influences directly the approximation error.
A suitable choice of this quantity enables an efficient smoothing of noisy data as shown in Figure \ref{MLS_classic}.
In our work the influence radius $D$ is chosen automatically.

In an earlier publication \cite{Roos_2007_WOST} a regularized 
weighting function was presented which enables the fulfillment of the MLS interpolation condition with high accuracy.
This approach is very suitable for problems where an interpolating meta-model is required. For problems of noisy input data
the noise is represented by the approximation function as shown in Figure \ref{MLS_reg} and thus the classical MLS approach with exponential weighting function is more suitable.
Due to the reduction of the variable space in the MOP approach, the occurring artificial noise would be represented if the interpolating MLS approach would
be used. For this reason we recommend only the use of the classical MLS method inside of the MOP framework.

The main advantage of the MLS approach compared to the polynomial regression is the possibility to represent 
arbitrary complex nonlinear (but still continuous) functions. By increasing the number of support points the approximation function 
will always converge to the exact formulation. Furthermore, the MLS approximation requires no computational demanding training algorithm as
Kriging or neural networks. This is a very important property required to accelerate the MOP algorithm.

\section{Adequacy of the regression model}
\subsection{Coefficient of determination}

In order to verify the approximation model, the 
Coefficient of determination $CoD$ has been introduced
\begin{equation}
CoD = 1 - \frac{\sum_{j=1}^n (y_j - \hat y_j)^2}{\sum_{j=1}^n(y_j - \bar y)^2}.
\label{COD}
\end{equation}
For polynomial regression models the coefficient of determination can be alternatively calculated as
\begin{equation}
CoD = \frac{\sum_{j=1}^n(\hat y_j - \bar y)^2}{\sum_{j=1}^n(y_j - \bar y)^2}, \text{ or } CoD = \rho^2_{\mathbf{y \hat y}}.
\end{equation}
For a perfect representation of the support point values this measure is equal to one.
This can be the case if the approximation can represent the model perfectly
or if the degree of the basis polynomial is equivalent to the number of support points.
In order to penalize the second case the adjusted coefficient of determination was introduced
\begin{equation}
CoD_{adj} = 1 - \frac{n-1}{n-p}(1-R^2),
\end{equation}
where $p$ is the number of coefficients used in the regression model.
\begin{figure}[th]
	\centerline{\includegraphics[width=0.6\textwidth]{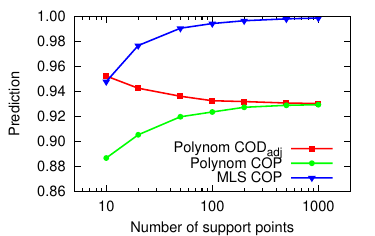}}
\caption{Convergence of coefficient of determination and the coefficient of prognosis
depending on the number of support points of a linear polynomial and MLS approximation
of a coupled nonlinear two-dimensional model}
\label{prediction}
\end{figure}
In Figure \ref{prediction} the convergence of the $CoD_{adj}$ is shown for a linear approximation
of a two-dimensional quadratic model ($Y=2 X_1 + 4 X_2 + 0.5 X_1^2 + X_1 X_2$) with standard normally distributed
input variables. The figure indicates, that for small numbers of support points the $CoD_{adj}$
is to optimistic and gives an overestimate of the approximation quality.

\subsection{Coefficient of prognosis}
In order to improve our quality measure we introduce the Coefficient of prognosis
based on cross validation.
In the cross validation procedure the set of support points is mapped to $q$ subsets. 
Then the surrogate model is built up by removing subset $i$ from the support points
and approximating the subset model output $\tilde y_j$ using the remaining point set.

The Coefficient of prognosis is now formulated 
with respect to the general definition of the $CoD$ according to Eq. \ref{COD}.
In contrast to the polynomial $CoD$ in Eq. \ref{COD}, this value can be slightly negative if the variance of the residuals is larger as 
the variance of the response itself. However,
this case indicates that the meta-model as not suitable for the approximation.
Other formulations for the $CoP$ as the 
squared
linear correlation coefficient between the approximated and true model outputs
are not suitable for all meta-model types.
In Figure \ref{prediction} the $CoP$ values are shown for the simple example using polynomial regression and MLS approximation.
The figure indicates, that the $CoP$ values are not to optimistic for small sample numbers as the $CoD$ values.

The evaluation of the cross validation subsets, which are usually between 5 and 10 sets, causes additional numerical effort 
in order to calculate the $CoP$. Nevertheless, for polynomial regression and Moving Least Squares, this additional effort
is still quite small since no complex training algorithm is required. For other meta-modeling approaches 
as neural networks, Kriging and even Support Vector Regression,
the time consuming training algorithm has to be performed for every subset combination,
which makes an application inside of the MOP framework not very attractive. 

\begin{figure}[p]
	\begin{minipage}[t]{.33\linewidth}
	\centerline{\includegraphics[width=1.1\textwidth]{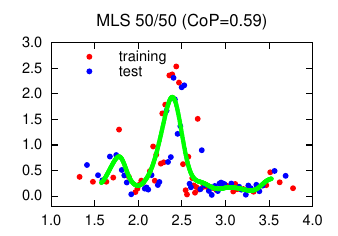}}
	\end{minipage}\hfill
	\begin{minipage}[t]{.33\linewidth}
	\centerline{\includegraphics[width=1.1\textwidth]{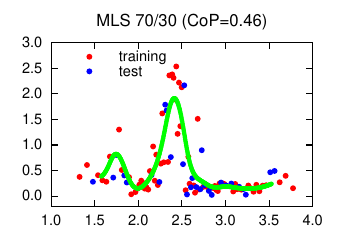}}
	\end{minipage}\hfill
	\begin{minipage}[t]{.33\linewidth}
	\centerline{\includegraphics[width=1.1\textwidth]{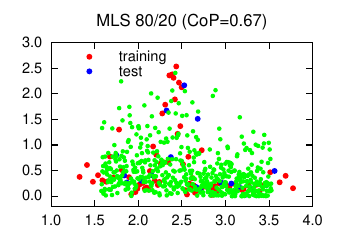}}
	\end{minipage}\\
	\begin{minipage}[t]{.33\linewidth}
	\centerline{\includegraphics[width=1.1\textwidth]{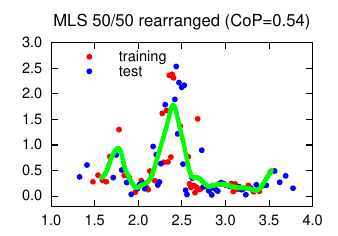}}
	\end{minipage}\hfill
	\begin{minipage}[t]{.33\linewidth}
	\centerline{\includegraphics[width=1.1\textwidth]{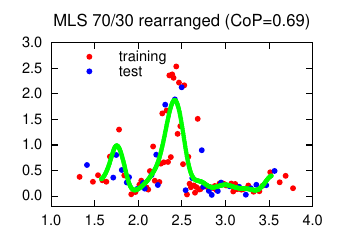}}
	\end{minipage}\hfill
	\begin{minipage}[t]{.33\linewidth}
	\centerline{\includegraphics[width=1.1\textwidth]{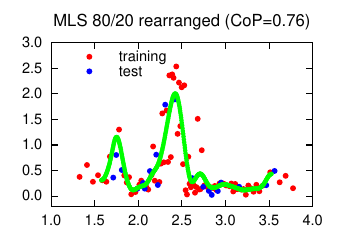}}
	\end{minipage}
  \caption{Approximation with corresponding $CoP$ values using data splitting}
  \label{COP_testdata}

\vspace{10mm}
	\begin{minipage}[t]{.33\linewidth}
	\centerline{\includegraphics[width=1.1\textwidth]{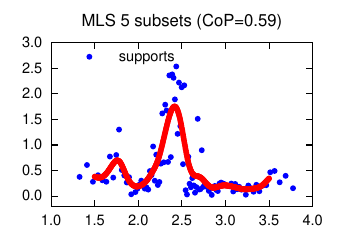}}
	\end{minipage}\hfill
	\begin{minipage}[t]{.33\linewidth}
	\centerline{\includegraphics[width=1.1\textwidth]{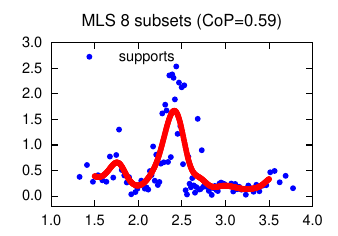}}
	\end{minipage}\hfill
	\begin{minipage}[t]{.33\linewidth}
	\centerline{\includegraphics[width=1.1\textwidth]{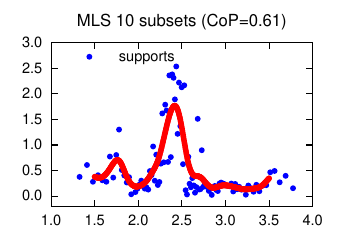}}
	\end{minipage}\\
	\begin{minipage}[t]{.33\linewidth}
	\centerline{\includegraphics[width=1.1\textwidth]{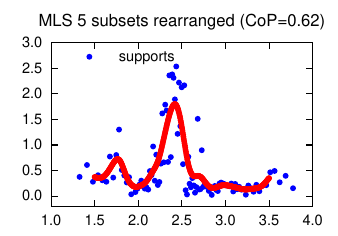}}
	\end{minipage}\hfill
	\begin{minipage}[t]{.33\linewidth}
	\centerline{\includegraphics[width=1.1\textwidth]{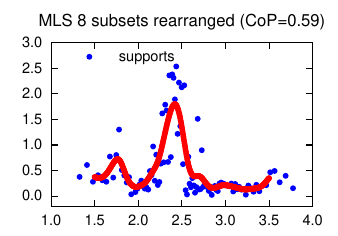}}
	\end{minipage}\hfill
	\begin{minipage}[t]{.33\linewidth}
	\centerline{\includegraphics[width=1.1\textwidth]{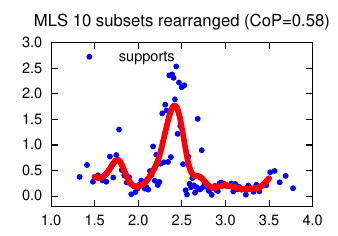}}
	\end{minipage}
  \caption{Approximation with corresponding $CoP$ values using cross validation}
  \label{COP_cross}
\end{figure}
In an earlier implementation \cite{Most_2008_WOST} we used a splitting of the original data 
in test and training points
to calculate the coefficient of prognosis. There, we could observe a strong dependence 
of the $CoP$ estimates an the way how the samples were split, although the approximation
was almost identical. This is shown in Figure \ref{COP_testdata}.
With the application of the cross validation procedure this problem could be solved and 
much more reliable estimates of the $CoP$ could be obtained as shown in Figure \ref{COP_cross}.

\section{Sensitivity of input parameters}
\subsection{Coefficient of Importance}
Generally the coefficient of determination $CoD$ is interpreted as the fraction of the variance of the true model
represented by the approximation model. This can be used to estimate
the Coefficient of Importance based on the regression model 
\begin{equation}
CoI_{i} = CoD_{\mathbf{X}} - CoD_{\mathbf{X}_{\sim i}},
\label{coi}
\end{equation}
where $CoD_{\mathbf{X}}$ is obtained using the complete parameter set to build up the regression model
and $CoD_{\mathbf{X}_{\sim i}}$ originates from a regression model with the reduced parameter set $\mathbf{X}_{\sim i}$.

\subsection{Coefficient of Prognosis for single variables}
\begin{figure}[th]
	\begin{minipage}[t]{.48\linewidth}
	\centerline{\includegraphics[width=1.15\textwidth]{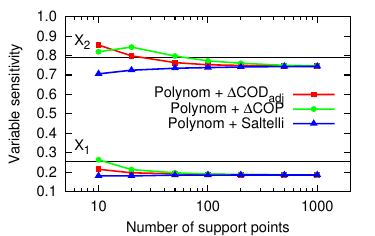}}
	\end{minipage}\hfill
	\begin{minipage}[t]{.48\linewidth}
	\centerline{\includegraphics[width=1.15\textwidth]{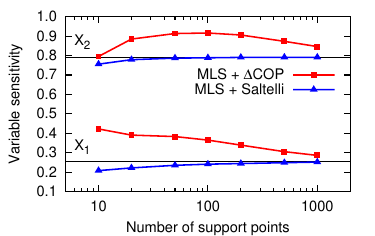}} 
	\end{minipage}
  	\caption{Convergence of the sensitivity indices for the nonlinear two-dimensional model}
   	\label{indeces}
\end{figure}

Analogous to the coefficient of importance, the $CoP$ for single variables 
can be defined as the difference between the full and the reduced approximation model.
In Figure \ref{indeces} these values are shown in comparison to the $CoI$ values using polynomials
and also using MLS approximation for the two-dimensional quadratic model. 
The figure indicates a strong overestimation of the influence of the single variables 
for the MLS approximation if the number of supports is small.

For this reason total effect sensitivity indices
are used instead. 
For our purpose these indices are directly evaluated on the regression model using the standard estimators
with a large number of samples. The 
resulting estimates of the indices are finally scaled with the $CoP$ of the regression model.
In Figure \ref{indeces} the estimated total sensitivity indices are shown additionally which
converge significantly faster than the estimates using the reduction approach.
For purely additive models the single variable indices sum up to the total $CoP$ of the regression model.
If coupling terms are present in the approximation, their sum is larger than the total $CoP$.

\begin{figure}[th]
	\centerline{\includegraphics[width=0.5\textwidth]{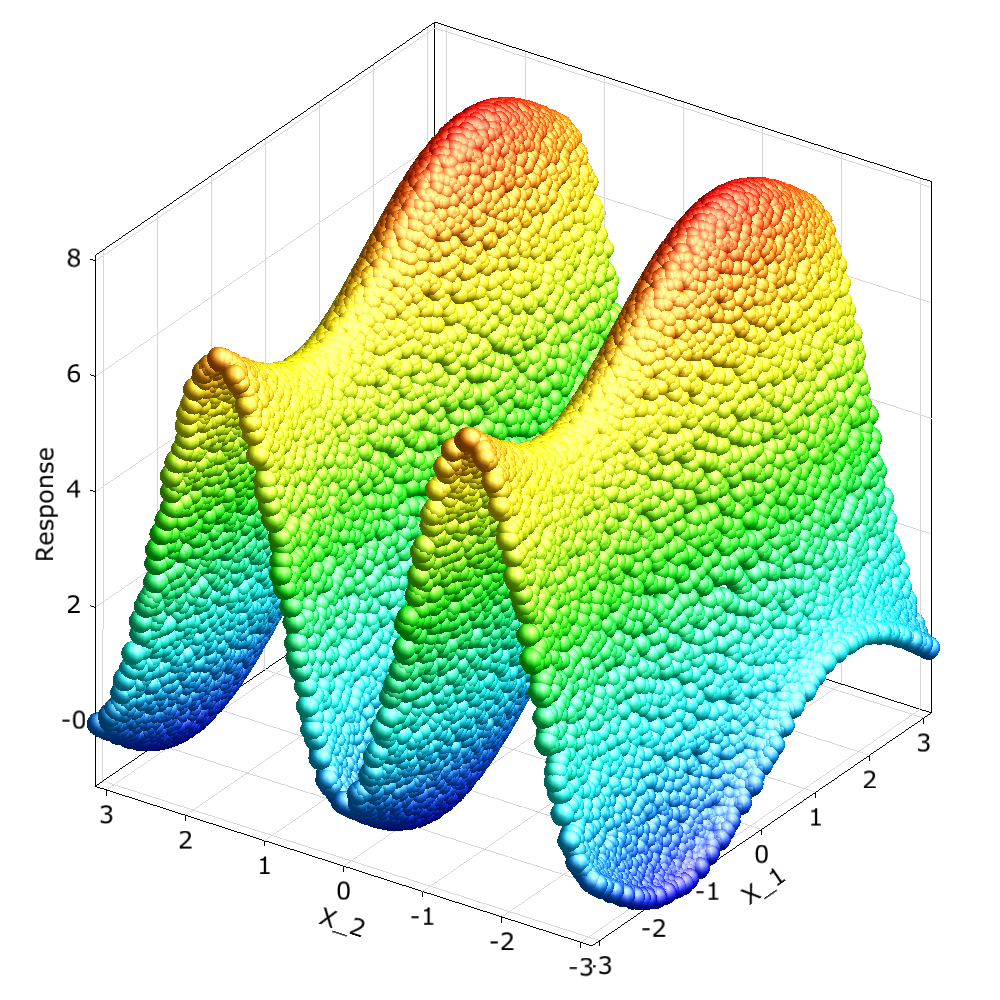}}
	\vspace{2mm}
\begin{tabular}{cccccc}
    \hline
	&Analytical & Splitting 50/50 & Splitting 70/30 & Splitting 80/20 & Cross validation \rule[0mm]{0pt}{2.5ex}\\
   \hline
	$CoP$	& -   & 0.95 & 0.98 & 0.99 & 0.96\rule[0mm]{0pt}{2.5ex}\\
	$CoP_1$	&0.56 & 0.57 & 0.61 & 0.61 & 0.58\rule[0mm]{0pt}{3.5ex}\\
	$CoP_2$	&0.44 & 0.38 & 0.39 & 0.40 & 0.40\\
	$CoP_3$	&0.24 & 0.27 & 0.26 & 0.26 & 0.25\\
   \hline
\end{tabular}
\caption{Ishigami test function and calculated sensitivity indices using 500 samples with 
data splitting and cross validation}
\label{ishigami}
\end{figure}

In Figure \ref{ishigami} the estimated sensitivity indices are given for the well known Ishigami function.
The figure indicates a very good agreement with the analytical values. As discussed above, the data splitting
$CoP$ may strongly depend on the splitting fractions. In this example the $CoP$ values are to optimistic 
for a small amount of test samples. This is not the case if cross validation is applied 
where the sensitivity indices
are more reliable.

\section{MOP framework}
\subsection{Significance filter}

\begin{figure}[th]
	\includegraphics[width=0.49\textwidth]{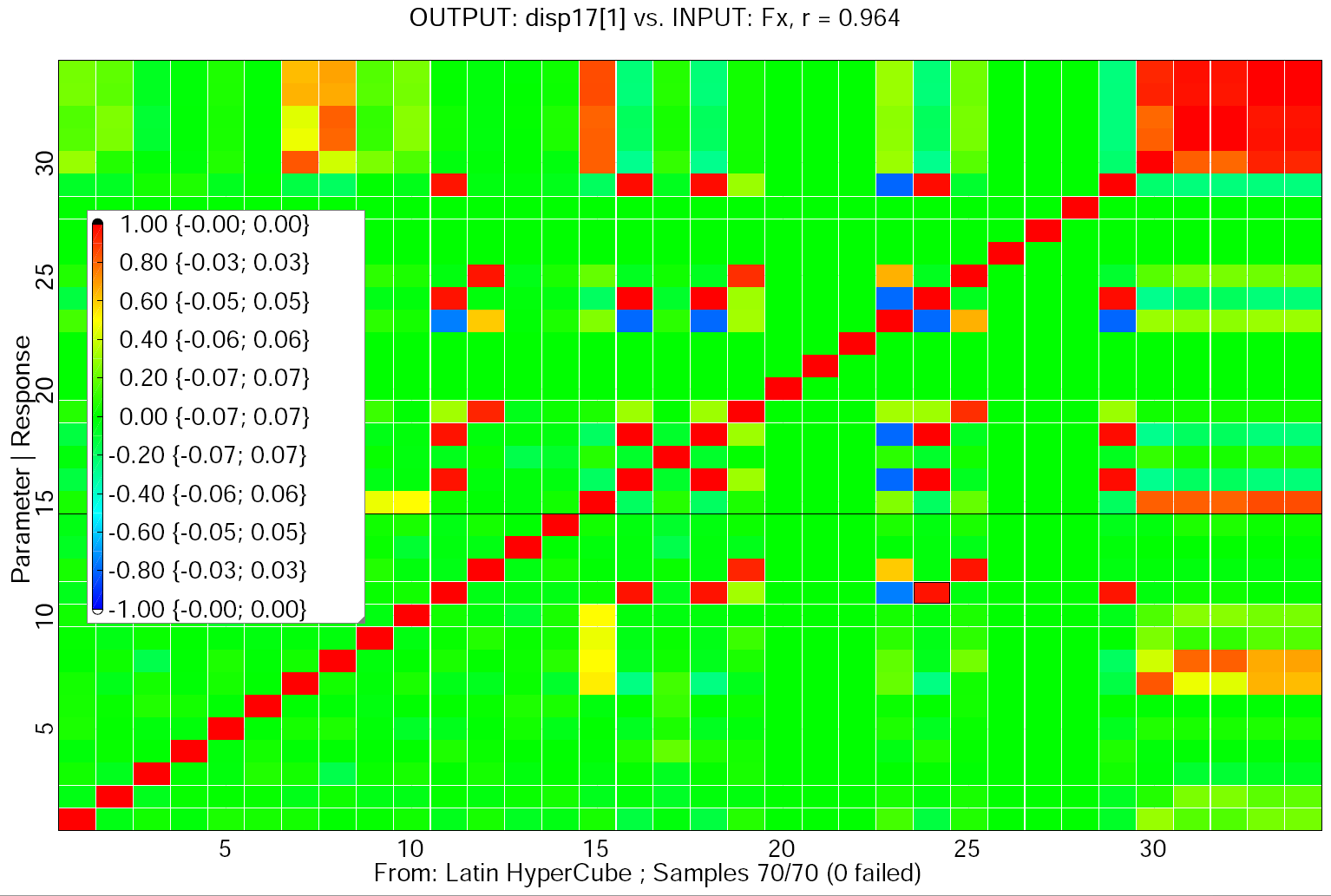}
	\hfill
	\includegraphics[width=0.49\textwidth]{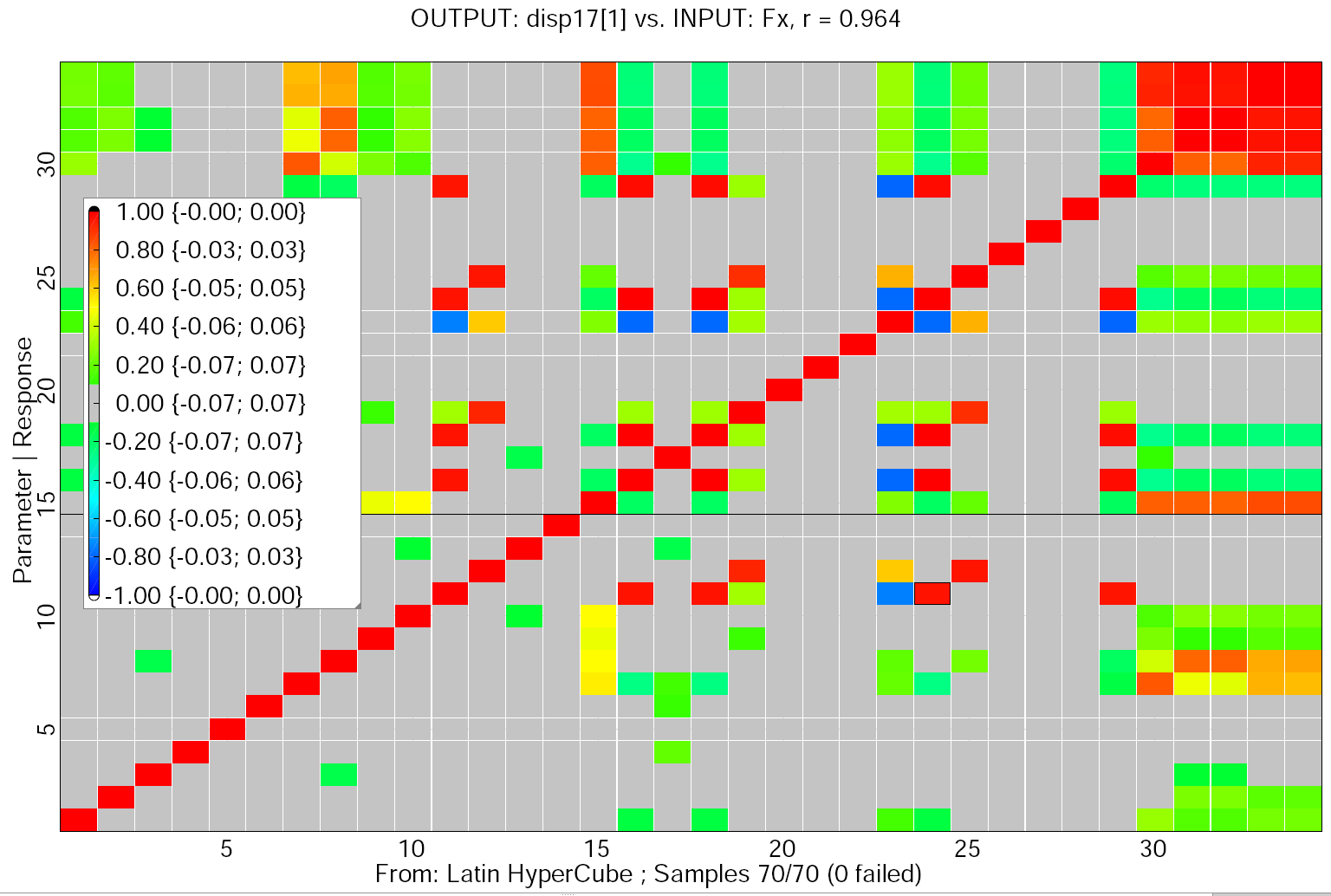}
  		\caption{Full linear correlation matrix and reduced matrix considering the most significance 
		linear correlation coefficients}
   		\label{lin-matrix-filter}
\end{figure}

The coefficients of correlation $\rho_{ik}$ are calculated from all 
pairwise combinations of both input variables and response according to:
\begin{equation}
\rho_{ij} =  \frac{1}
{{n - 1}}\quad \frac{\sum\limits_{k = 1}^N {\left( {x_i^{(k)}  - \mu_{x_i} } \right)
\left( {x_j^{(k)}  - \mu_{x_j} } \right) }}{\sigma_{x_i}\sigma_{x_j}}.
\label{coefficientscorrelation}
\end{equation}
The quantity $\rho_{ij}$, called the linear correlation coefficient, measures the strength and
      the direction of a linear relationship between two variables. 
All 
pairwise combinations ($i,j$) can be 
assembled 
into a correlation matrix as shown in Figure \ref{lin-matrix-filter}.

The computed correlation coefficients between the input variables vary from the assumed values depending on the sample size.
This deviation is used to judge in a first step, which variables are significant concerning there influence on the output variables.
We define an error quantile which is chosen between $90 \%$ and $99 \%$ and compute the corresponding correlation error
in the input-input correlations. This is done for linear and quadratic correlations simultaneously. Based on these quantile values
we assume only these input variables to be significant concerning an output variable if there correlation values are above the given error
values. For the most practical cases this leads to a reduced number of input variables which is shown in Figure \ref{lin-matrix-filter}.
All values in gray are assumed to be insignificant.

\subsection{Importance filter}
Generally the remaining variable set still contains variables which are not needed for an approximation.
With the importance filter we identify the important variables for the approximation model as described as follows:
Based on a polynomial regression using the remaining variables of the significance filter we
estimate the quality of the model representation by the coefficient of determination ($CoD$).
Based on a given value of the minimum required coefficient of importance $CoI_{min}$ only the variables having 
\begin{equation}
CoI_i \geq CoI_{min}
\end{equation} 
are considered in the final approximation. Generally the value $CoI_{min}$ is taken between $1 \%$ and $9 \%$.

\subsection{Determination of the optimal meta-model}
Based on the definition of the coefficient of prognosis we can derive the optimal meta-model with corresponding variable space as
follows:
For each meta-model type we investigate all possible significance and filter configurations by varying the significance quantile from
$99 \%$ down to a given minimal value. Then a polynomial regression is built up and the coefficients of importance are calculated for
each variable. The threshold $CoI_{min}$ is varied from $0.01$ to a given value and based on the remaining variables the
meta-model is built up and the coefficient of prognosis is computed.
The configuration with the maximum $CoP$ based on cross validation is finally taken as optimal meta-model with corresponding variable space for each
approximated response quantity. 

Finally a certain reduction of the best $CoP$ by a given $\Delta CoP$ around $3\%$ is allowed in order reduced further the number of 
input variables and even the complexity of the meta-model.
This means, that if for example MLS approximations gives the best $CoP$ of $85 \%$, but polynomial regression with the same number of inputs
leads to $83 \%$, polynomial regression, which is the simpler model, is preferred.
This helps the designer to explain the response variation  with a minimum number of inputs while keeping the approximation model
as simple as possible.

\subsection{Implementation of CoP/MOP in optiSLang}
Since version 3.1 the CoP/MOP approach is included in the commercial software 
{optiSLang} \cite{optiSLang2009}. There is a flow created which can evaluate CoP/MOP for any set of samples 
or test data. The user can choose cross validation or data splitting, 
the different filter settings and can force to reduce (Delta CoP) the number of important variables 
in the final MOP model.

\begin{figure}[th]
	\includegraphics[width=\textwidth]{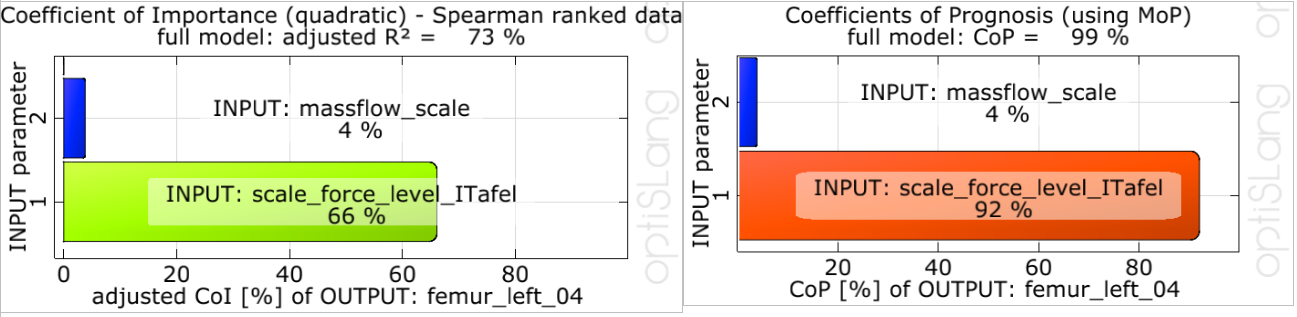}
  		\caption{Left: Coefficient of Importance using traditional correlation analysis, right: Coefficient of Prognosis using CoP/MOP approach}
   		\label{cop_plot}
\end{figure}

\begin{figure}[th]
\centering
	\includegraphics[width=0.5\textwidth]{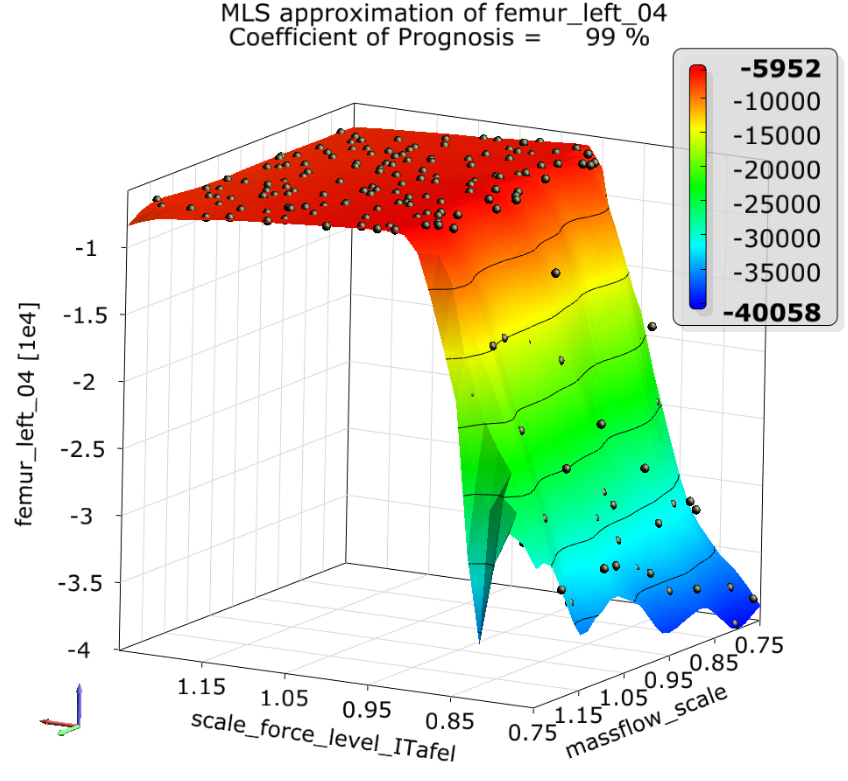}
  		\caption{Visualization of the MOP approximation in the subspace of the two most important variables}
   		\label{mop_plot}
\end{figure}

Since it is available in optiSLang the generation of the meta-model of optimized prognosis and the 
calculation of coefficient of prognosis was successfully applied at several problem types. 
The following example shows a noise non-linear problem having 8 optimization variables and 
200 samples. Running traditional correlation analysis using Spearman ranked data two important 
variables could be identified and a $CoI$ of $73\%$ (Fig. \ref{cop_plot}) for the full model was measured. 
Running CoP/MOP we also find the two important variables, but with very good representation 
of the nonlinear response function we can achieve a $CoP$ of $99\%$ (Fig. \ref{cop_plot},\ref{mop_plot}).   

From our experience so far we can state and recommend:
\begin{itemize}
\item Compare $CoI$ of correlation analysis and $CoP$, the differences between the two should be verified.
\item Check plausibility and prognosis quality by plotting the MOP with the two most important input variables.
\item If the MOP approach can reduce the set of important to a very small number ($<5$), 
very good representation of nonlinearities are achieved even with small number of samples (50).
\item If the MOP approach cannot reduce the set of important input parameters smaller than $10\dots 20$, 
the sample set must be large to represent non-linear correlation.
\item The $CoP$ measure of the full model is more reliable than the $CoI$ measure of the full model. 
\end{itemize}

\section{Conclusion}
In this paper we presented an approach for an automatic selection of the optimal meta-model for a specific problem.
We introduced the coefficient of prognosis which enables an objective assessment of the meta-model prognosis based on cross validation.
The approach can identify the required variables efficiently and the obtained optimal meta-model can be used afterwards for an optimization. 
We could show that cross validation gives more reliable estimates of the $CoP$ for small samples sizes than sample splitting with
only minimal additional numerical costs for polynomials and MLS.
Furthermore, Saltelli's approach for sensitivity analysis combined with MOP could improve the accuracy of the variable indices significantly.

In our future work we will address the following improvements:
\begin{itemize}
\item Consideration of dependent inputs in the sensitivity measures.
\item Development of filter procedures for the detection of coupling terms.
\item Bootstrapping of the $CoP$ in order to assess its estimation quality depending on the number of samples.
\end{itemize}
After generation of MOP the next step will be to use the MOP for optimization purpose within optiSLang. 
Then a black box algorithm combining high end sensitivity study and optimization will be available 
for medium and high dimensional non linear problems.


\end{document}